# Customized video filtering on Youtube

Vishal Anand[1], Ravi Shukla[2] , Ashwani Gupta[3] , Abhishek Kumar[4]

[1] Department of Mechanical Engineering
Indian Institute of Technology Kharagpur, Kharagpur India
vishal.anand456@iitkgp.ac.in
[2] Department of Computer Science and Engineering,
Indian Statistical Institute, Kolkata
ravichamp2020@gmail.com
[3] Senior Data Scientist, Publicis Sapient
ashwani.gupta@publicissapient.com
[4] Senior Manager, Publicis Sapient
abhishek.kumar@publicissapient.com

**Abstract.** Inappropriate and profane content on social media is exponentially increasing and big corporations are becoming more aware of the type of content on which they are advertising and how it may affect their brand reputation. But with a huge surge in content being posted online it becomes seemingly difficult to filter out related videos on which they can run their ads without compromising brand name. Advertising on youtube videos generates a huge amount of revenue for corporations. It becomes increasingly important for such corporations to advertise on only the videos that don't hurt the feelings, community or harmony of the audience at large.
In this paper, we propose a system to identify inappropriate content on YouTube and leverage it to perform a first of its kind, large scale, quantitative characterization that reveals some of the risks of YouTube ads consumption on inappropriate videos. Customization of the architecture have also been included to serve different requirements of corporations. Our analysis reveals that YouTube is still plagued by such disturbing videos and its currently deployed countermeasures are ineffective in terms of detecting them in a timely manner. Our framework tries to fill this gap by providing a handy, add on solution to filter the videos and help corporations and companies to push ads on the platform without worrying about the content on which the ads are displayed.

**Keywords:** Youtube, Inappropriate video filtering, NLP, Feature Extraction, Deep Learning, Customisation , Flask API, HAN

# 1 Introduction

Social media allows people, companies and organisations to exchange or share information such as pictures, videos, career interests, ideas, philosophy, lessons etc. to virtual communities and networks. Websites like Youtube, Facebook, Linkedin, Pinterest, etc provide effective platform for communication among people around the world. On the other hand, misuse of such platforms to insult, abuse, harass or target others are not uncommon since these websites are used by the public at large and the information conveyed can be disturbing or controversial. Among the other social media tools, video content has far more outreach and has long lasting impact. Recent studies have shown that Youtube, Vimeo, Twitch and Daily-motion are one of the most popular video sharing websites today. Among them youtube has the largest audience base with around 2 Billion monthly active users with more than 500 hours of video uploaded every minute. Such amount of custom filtering of videos manually is impossible and is prone to human errors and also has language barriers.

Companies and organisation always try to protect their brand integrity and don't want to mislead their audience or promote something which can affect their brand status. Its violation may easily happen if their advertisements runs on content which are severely inappropriate and profane. If proper filtering and checking is not done, it proves to be devastating to companies brand reputation. In 2018, a CNN investigation has found that ads from over 300 companies and organisations which includes tech giants, major retailers, newspapers, and government agencies ran on Youtube channels promoting white nationalists, Nazis, pedophilia, conspiracy theories, and North Korea propaganda. Even ads from US government agencies, such as the Department of Transportation and the Centers for Disease Control, also appeared on inappropriate channels. Companies pays attention to where and how their brands are represented, to ensure alignment with their strong companies values. They work very hard to target the right customers and ensure that their ads appear within appropriate content. But unavailability of a robust and reliable filter mechanism can lead to wastage of millions of dollars.

Offensive videos create a big impact on a large reach of people. Youtube has a set of guidelines aimed to reduce the inappropriate videos on their website. Despite these guidelines there are significant proportions of videos that have unwanted or offensive content. There is no appropriate filter to check the contents of video on which one can run ads on. This unavailability of a screening mechanism to filter out video channels which may insult some religion, criticize race, make sexist comments, make fun of disability, promote extremist or terrorist content may defame a brands reputation if ads are run on such channels. It has become very challenging for companies and organisation to run advertisements considering the fact that there is a large pool of such profane content on youtube.

Companies handle this situation by hiring people for manual filtering of youtube channels which have offensive or profane content. This practise suffers from various flaws like additional expense, time consuming, prone to human errors and language barriers.

In our work we have tried to solve this problem by leveraging the power of machine learning and AI. We have provided a custom filter which can be used by organisations to filter out only selected type of inappropriate content based on their core values and ethics. This flexibility allows anyone to use the framework suited to their needs and desires. Additionally there is a threshold to filter strictness which help to decide what level of a particular profanity can be tolerable.

The key components of our proposed customised filtering system are:

- Collection of youtube data through scrapers and api
- Building of a labelled dataset for classification
- Extraction of audio visual features
- NLP of textual features and processing of video features
- Building Classification model and comparing their performance
- Designing the framework into an API

We have finally wrapped up our engineering into a working API which could instantaneously provide results showing inappropriate content of a youtube channel on a scale of 0 to 100%. The user needs to input two files. One containing the offensive word dictionary which contains the list of words which the user dont want on the youtube video on which ads are run. Second is a list of youtube channel list which contains all the youtube channels which the user wants to filter out from. Having fed these two files to the API, would result in an inappropriateness scale from 0-100% for each and every video channel in the list. The user can then manually set a threshold of the strictness which can be tolerable for running ads and filtering those videos out of the rest.

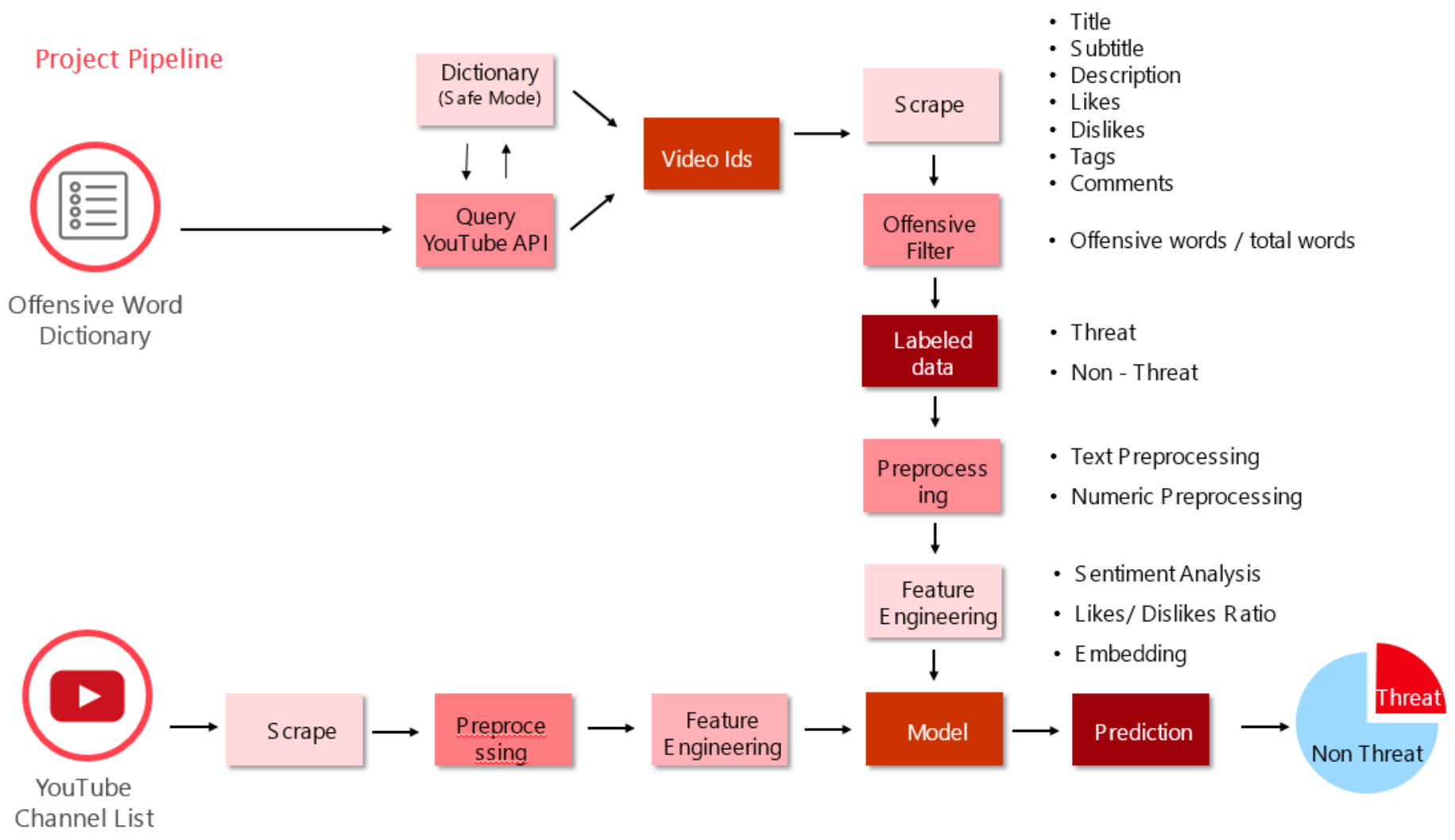

Figure 1: Pipeline for customised video filtering on youtube.

## 2 Literature Survey

Chen et al. [1] introduced the idea of identifying offensive language in social media. They proposed a lexical syntactic architecture which incorporates both message-level and user-level features to identify offensive messages and offensive users. It is based on the idea that in a sentence if profanities are associated with user or contain other offensive words, then the sentence is offensive. For this purpose, this approach uses two dictionaries, i.e., strong and weak offensive word dictionaries. For instance, f**k and s**k come under strong profanities, stupid and liar are weak profanities. The architecture assigns different weights for strong and weak offensive words. It uses Stanford dependency parser to identify association of offensive word/s or user identifier to offensive word. Dinakar et al. [2] proposed an approach to detect textual cyberbullying. This approach first uses binary classifier to identify sensitive text. After spotting sensitive text, this framework builds topic sensitive classifiers by analyzing textual comments posted on a topic. This approach makes use of a list of profane words, TF-IDF, lexicon of negative words and label-specific unigrams, bigrams as features. Dadvar et al. [3] improved detection of cyberbullying with user information. They collected top three videos from different video categories such as entertainment, politics and sports. This framework comprises of three sets of features, namely, content-based features, cyber-bullying features and user-based features. Reynolds et al. [4] proposed an approach to detect cyberbullying using machine learning. It discovers language patterns used by bullies and victims based on hard coded rules. The idea of this work is to focus on number of bad words along with their intensity used in the video. Agarwal et al. [5] focused on detecting privacy invading harassment and misdemeanor videos. Feature set of their approach includes linguistic (percentage of keywords in title and description), popularity (ratio of likes and views, ratio of comments and views), duration (duration of video) and category (YouTube category). It uses one class classifier to detect objectionable videos. Rafiq et al. [6] designed a technique to detect cyberbullying instances in Vine1 , a mobile based video-sharing social network that allows users to record and edit six-second videos. Their work distinguishes cyberbullying and cyber aggression. It uses CrowdFlower2 , a crowd-sourced website to annotate media sessions for cyberbullying and cyber aggression. Fu et. al [7] proposed an approach to identify extremist videos in online video sharing sites. This work uses user-generated content such as comments to identify extremist groups. Feature set of this work includes lexical, syntactic and content-specific features to identify extremist groups. SVM with 10-fold cross validation was adopted to classify extremist videos.

In summary, the existing work has been done on classifying videos based on comments and user profiles but only a limited research has been based on the video based features. In our approach we have tried to bridge this research gap by using video based features like subtitles, description, title, number of views, likes, dislikes, comments, images and video frames. This methodology not only gives relevant information about the video itself but at the same time reduces the error types in classification due to varied division of features.

# 3 Methodology

## 3.1 Building offensive word dictionary

For initial training and processing we compiled a list of 400+ words and phrases which includes a vast collection of inappropriate and profane vocabulary which could possibly be unaligned with the code of ethics of a company. Big corporations may not be willing to run ads on videos promoting such content or containing such words or phrases. We tried to add as many offensive words ranging from abuses, racist, sexist, nazi, terror and harassment. These compilations served as our initial customisation for filtering out such videos containing any relevance of these words in their content. We also built separate vocabulary of words which were targeted to one particular genre of offensiveness for ease for the users to choose their own set of vocabulary for filtering thereby allowing customisation from users side. Companies and brands may add or create their own vocabulary for filtering out videos based on their requirement and brand values they align to.

## 3.2 Dataset generation

Having the vocabulary at hand the next task involved making a collection of youtube videos which had offensive and non offensive content. This way machine learning model differentiate between offensive content and non offensive content effectively. We used Youtube API to return the search results of videos related to a particular word or phrase sent to YouTube API. All the compiled list of offensive words were sent to YouTube API to return related videos containing such content. Having done this we were able to accumulate around 8000 video ids of youtube videos which had high possibility of containing offensive content. Thereafter a similar procedure was carried out to collect video ids of videos having decent content. It was to make sure that the videos were collected from diverse backgrounds and content in order to maintain homogeneity and remove bias occurring due to non diversification. The procedure accumulated 20000+ video ids containing offensive and non offensive content which would be used for further process in pipeline.

## 3.3 Scraper to extract video features

The video ids generated by youtube API are used to scrap out important information from the videos. YouTube videos contains a lot of useful data.

- Title
- Subtitle
- Description
- Likes
- Dislikes
- Comments
- Views
- Subscribers
- Thumbnail

We used beautiful soup, scrapy and a pypi package - youtube-transcript-api for scraping out data from each and every youtube video present in our 20k video collection. Scrapy was used to scrap Title, Description, Likes, Dislikes, Views, Subscribers and thumbnails. Beautiful Soup was used to crawl the youtube for extraction out comments from each and every videos, while youtube-transcript-api was used to collect transcripts/subtitle of the videos. Data collection was a time taking process. At the end of the web crawling and scraping we had a huge dataset containing all the features of the videos. The features contained numeric and textual data. Additionally we also scrapped out compressed thumbnails of each youtube videos and stored them separately. The metadata of youtube video can be seen in figure 2.

We also incorporated frame level and video level features in the dataset which contained metadata of the video content itself. This data has raw information of what is the actual content of each video. We used the feature extractor provided in youtube-8m repository on github. This feature extractor is built by youtube itself to extract video and image level features from any video. The raw extracted features were in tfrecords format which were converted to dataframe and append to existing feature set of each video ids. The video level features had 1080 independent features while the image level features had 256 features.

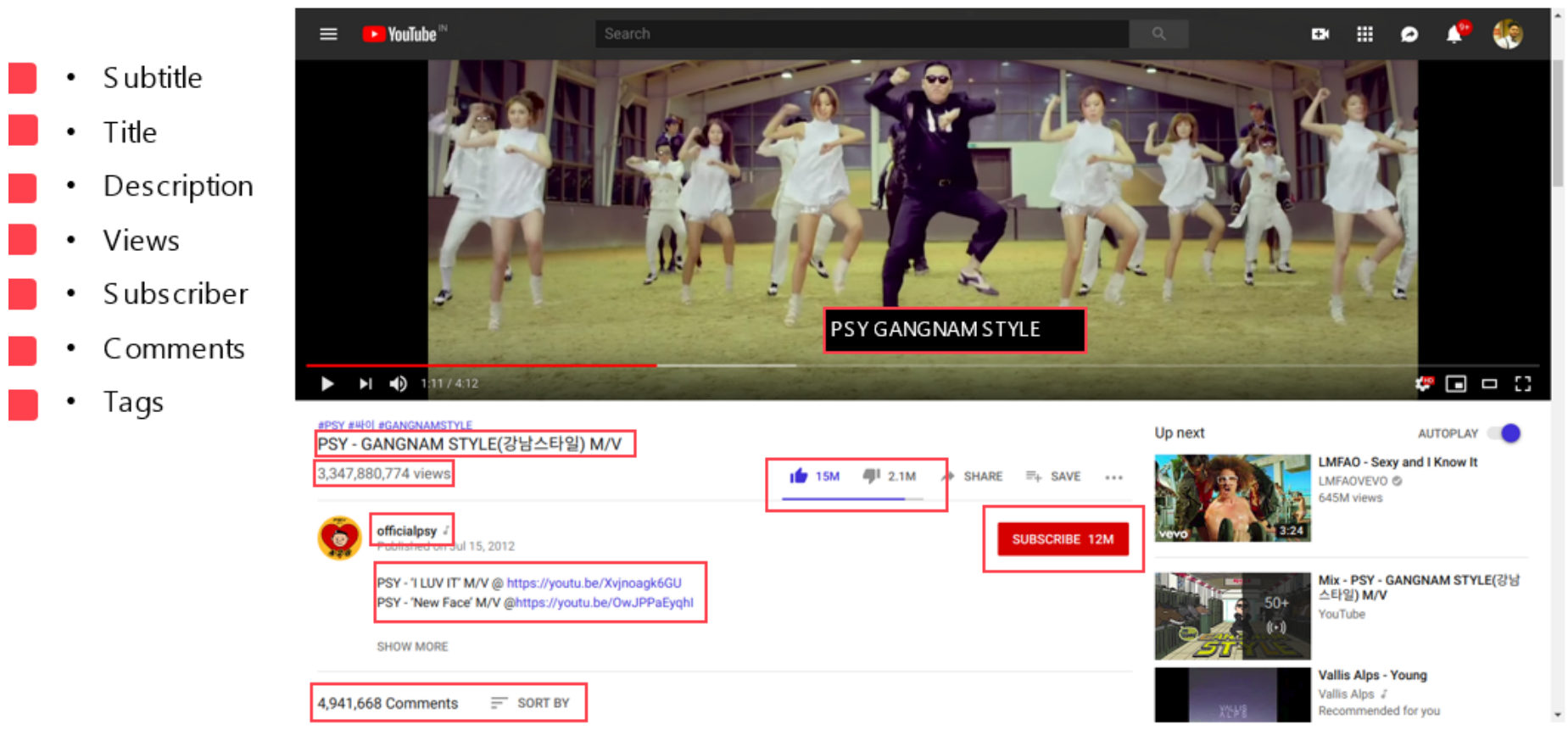

Figure 2: Youtube metadata extracted for building the dataset

This resulted in a dataset containing numeric, textual and image features ( thumbnails), providing a diverse feature set for properly classifying the videos based on the content they contain.

### 3.4 Offensive Filter and labelling

Having the entire videos data in place it is now necessary to label the videos as offensive or non offensive based on the offensiveness. We built an offensive filter which would use the textual data extract from videos for the purpose of filtering and labelling them for initial classification. The offensive filter use a counter for counting the number of offensive words or phrases that would occur in the entire text data. Based on this counter, it checks what percentage of these words from the entire text data. This text data includes title, subtitle and description. These three feature fields were merged to form one single text field for the labelling. If the percentage of offensive words/phrases in the entire text related to a video is more than a predefined threshold, the filter would label them as offensive or else non-offensive. Here we add this strictness parameter - threshold, to allow the users to decide on their own what level of tolerance they could bear with the videos while running their ads on them. This way users can control how much margin they can give to the filtered videos to have profane content.

We applied this approach because we wanted to label these videos and for supervised learning we needed to start with certain labelled data. In future work a better offensive filter could be built using the services like Amazon Mechanical Turk. Amazon MTurk enables companies to harness the collective intelligence, skills, and insights from a global workforce to streamline business processes, augment data collection and analysis, and accelerate machine learning development. The workforce can be asked to manually watch videos and label them as inappropriate or appropriate for better model building and classification. The labelling ended up with labelling more than 9000+ videos as profane and 11000+ videos as non-profane.

### 3.5 Preprocessing and feature Engineering

The youtube dataset contains numeric and textual features including thumbnail images of the video. This data needs to be preprocessed for inputting into the machine learning framework. First we would discuss the preprocessing done on textual data. In textual data we have Title, Description and Subtitle. We combined textual data to form one single long line of text and applied Natural Language Processing to clean the data and transform in a form ready for modelling.This preprocessing involves removal of stop words, tokenization, lemmatization and removal of special symbols from the text-based feature. Tokenization is the process of breaking a sentence into meaningful elements called tokens. Tokens that are very common but that do not contribute to significant relevant content are called stop words which need to be eliminated/filtered. After tokenization, we provide a standard stop-word list to remove such words from token list. Similarly, we remove unnecessary special symbols. We used regular expression pattern to capture special symbols to be eliminated. We used stop words list from Onix Text Retrieval Toolkit.

After the NLP we converted the text based features into embeddings. We used TF-IDF and pre trained word embeddings to transform the features to 2D Vectors representing the data. We used state of the art GloVe , fastText, and Word2Vec pretrained embeddings for textual representation in the form of vectors. GloVe is an unsupervised learning algorithm for obtaining vector representations for words. Pretrained embedding was trained on aggregated global word-word co-occurrence statistics from wikipedia corpus, and the resulting representations showcased interesting linear substructures of the word vector space. While fastText is a library for learning of word embeddings and text classification created by Facebook's AI Research lab. The model allows to create an unsupervised learning or supervised learning algorithm for obtaining vector representations for words. And Word2vec is a two-layer neural net that processes text. Its input is a text corpus and its output is a set of vectors: feature vectors for words in the corpus. While Word2vec is not a deep neural network, it turns text into a numerical form that deep nets can understand.

We have used all the embeddings which were pre trained on Wikipedia corpus. The pretrained embeddings solved the issue of training them on our own dataset which would have been computationally expensive. We used all three state of the art embeddings to convert the text based features into an n dimensional vectors for feeding our machine learning algorithms for training.

Having done this we had numbers based features like Likes, Dislikes, Numbers of subscribers, Views, image and video features. Handling Missing values and outlier detection was done to clean the data. The features were normalised and were transformed so that they follow normal distribution. Box Cox transformation was done to transform them into gaussian distribution. Features were numeric in nature and we used PCA to reduce the dimensionality of the dataset. We reduced the features to 100. We also had thumbnails of the videos. We converted the thumbnails to grayscale and extracted out pixel values of the 25X25 image. The image data was normalised and was kept aside for feeding into Deep Learning Framework.

Thus we had three different set of video features- 1) text based features 2) Numeric, image and video based features 3) Thumbnails based features. These three different feature sets followed different model pipeline whose results were finally merged for better prediction of the inappropriate and appropriate content.

### 3.6 Sentiment Analysis

Sentiment Analysis is the process of 'computationally' determining whether a piece of writing is positive, negative or neutral. It's also known as opinion mining, deriving the opinion or attitude of a speaker. We believed that comments on youtube video is also a great information which conveys how the video appealed to the audience and how did they react to it. We used a publically available twitter dataset which had comments from twitter post. The dataset had target variable as sentiment - whether the sentiment is harsh or decent. It gave a score between 0 and 1. 0 for decent and 1 for harsh. We trained a Beroulli Naive Bayes model to predict the sentiment of the tweets and used the trained model to predict the sentiment of youtube comments. We averaged out the score for 20 comments per video and gave a score to each video based on sentiment analysis. This score featured as a separate feature in the feature set.

### 3.7 Modelling

After all the preprocessing and feature extraction we built a baseline model using simple logistic regression. We flattened the 2D vectors which were used to form embeddings and merged all the features and inputted into the classifier. The classifier predicted the videos as a 0 or 1 , 0 being non offensive and 1 being offensive. The accuracy obtained using the baseline model was 62.87%

We used a different pipeline for modelling different types of features. We used LSTM with Hierarchical Attention Network on the 2D embedding features, CNN on the thumbnails features and a shallow neural network on numeric based features. Finally these deep neural networks were merged together and were flattened to pass onto fully connected layer followed by a softmax layer for classification. The best accuracy was attained by this architecture. Accuracy levelled up to 86% in classification. Some of the other machine learning algorithm used with different combinations of embeddings are shown in Table 1.

Table 1: Model Accuracy and embedding used

| Model + CNN+ Shallow NN | Embedding | Accuracy(%) |
|---|---|---|
| Logistic Regression | TF-IDF | 62.87 |
| SVM | TF-IDF | 67.70 |
| Convolution NN | Glove | 82.87 |
| LSTM | FastText | 81.33 |
| GRU | FastText | 84.77 |
| GRU+Hierarchical Attention Network | Glove | 86.4 |

## 3.8 Prediction Pipeline

Having built the classification pipeline, it was necessary for us to make a prediction pipeline for classifying a new unseen video that is encountered. The prediction pipeline follows a simple architecture. With the youtube channels provided by a user, we had our scraper in place which would use scrapy, beautiful-soup and youtube-transcript-api to fetch all the information from youtube videos in all the channels. The feature set was similar to the one we used for classification purpose. This scrapped data would then be preprocessed and feature engineering would be done on them as discussed in 4. This was followed by using the best model for prediction. Our prediction model used a LSTM architecture embedded with Hierarchical Attention networks in combination with CNN for image and shallow neural net for numeric based features. This model predicted whether the video is inappropriate or appropriate. Based on this classification we calculated what percentage of videos of a particular channel is inappropriate for running ads. The pipeline for prediction is shown in figure 3.

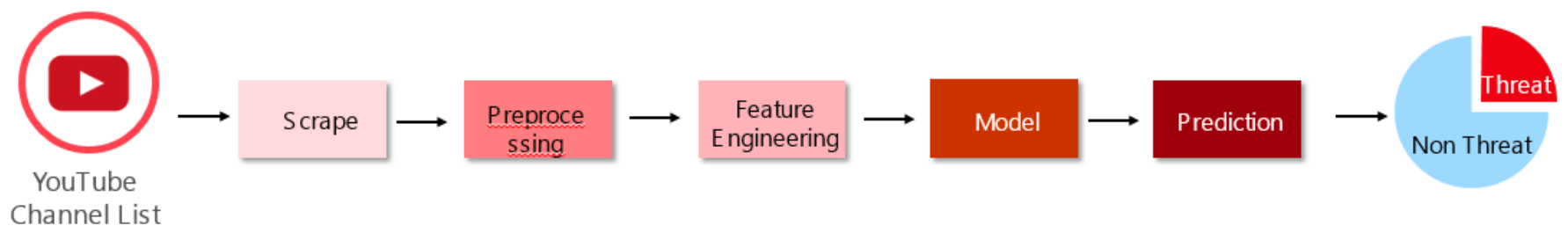

Figure 3: Prediction architecture for classifying into threat and non threat.

## 3.9 Flask API

After building the classification and prediction pipelines, we wrapped up the whole architecture in an easy to use API. We built an api on flask which would run on local server and would present a clean interface to users to interact with. The Flask API had three main tasks.

1) Getting the inputs
2) Backend classification
3) Presenting results and statistics

The user interface took two inputs- an offensive vocabulary as a text file and a file containing youtube channels urls. Submitting these two inputs to the api would result in activating the prediction pipeline. It would scrap out the video data of all the videos channels inputted followed by preprocessing the feature set into the format required for classification. The clean and processed data is fed into the ensembled deep learning architecture which would return back to the user interface with percentages of inappropriate video in each youtube channel. The threat detection UI is shown in figure 4. Addition to this we added some statistical graphs and visualisation of each youtube channel, which would depict the visualisation of the statistical features in each channel. The statistics included- likes/dislikes ratio, number of views, number of subs, video having maximum inappropriate content, videos having least profane content, wordgram of inappropriate texts found in the channel. A snapshot of the threat stats page is shown in figure 5.

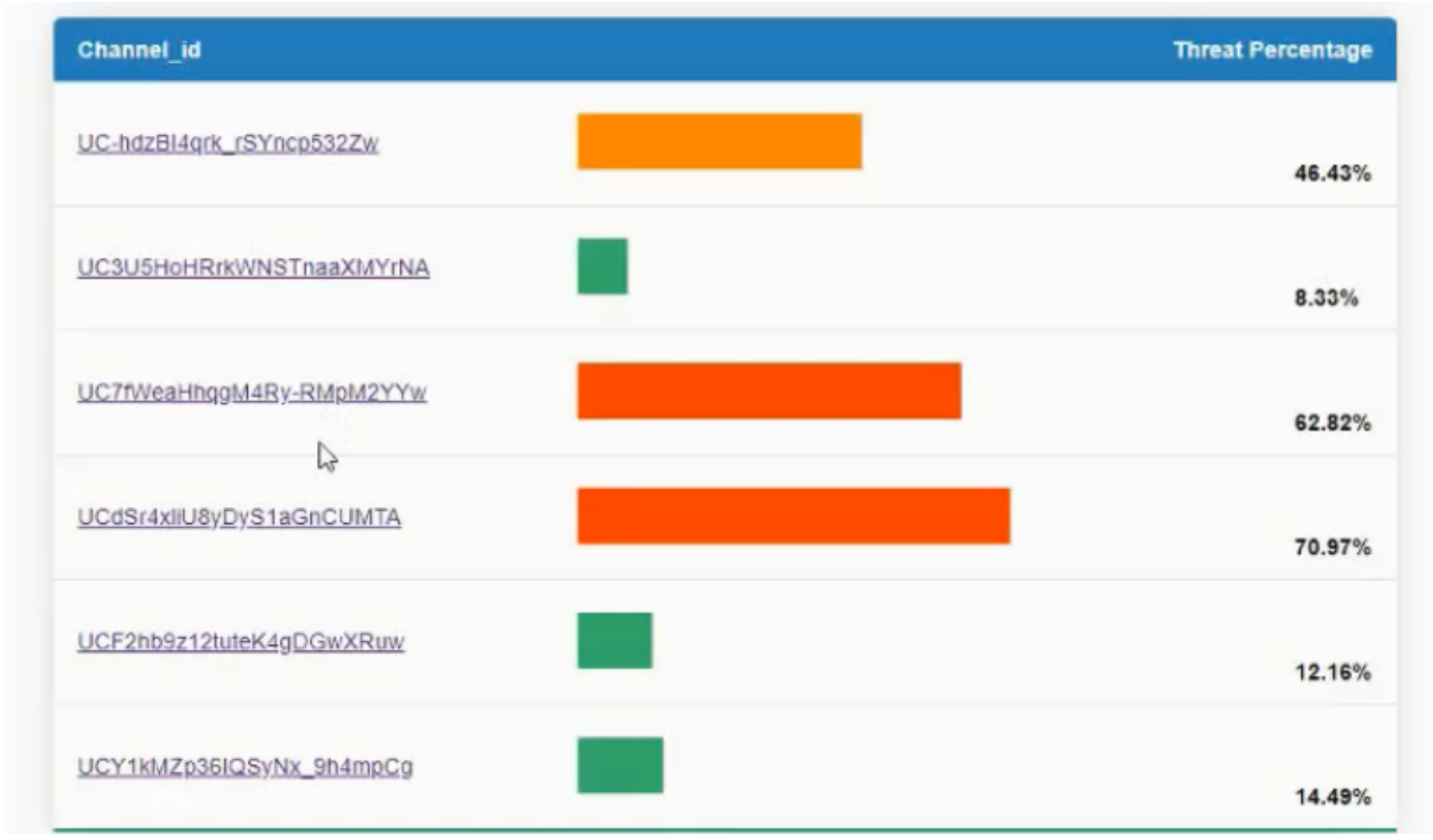

Figure 4: Sample threat percentages of videos in a youtube channel.

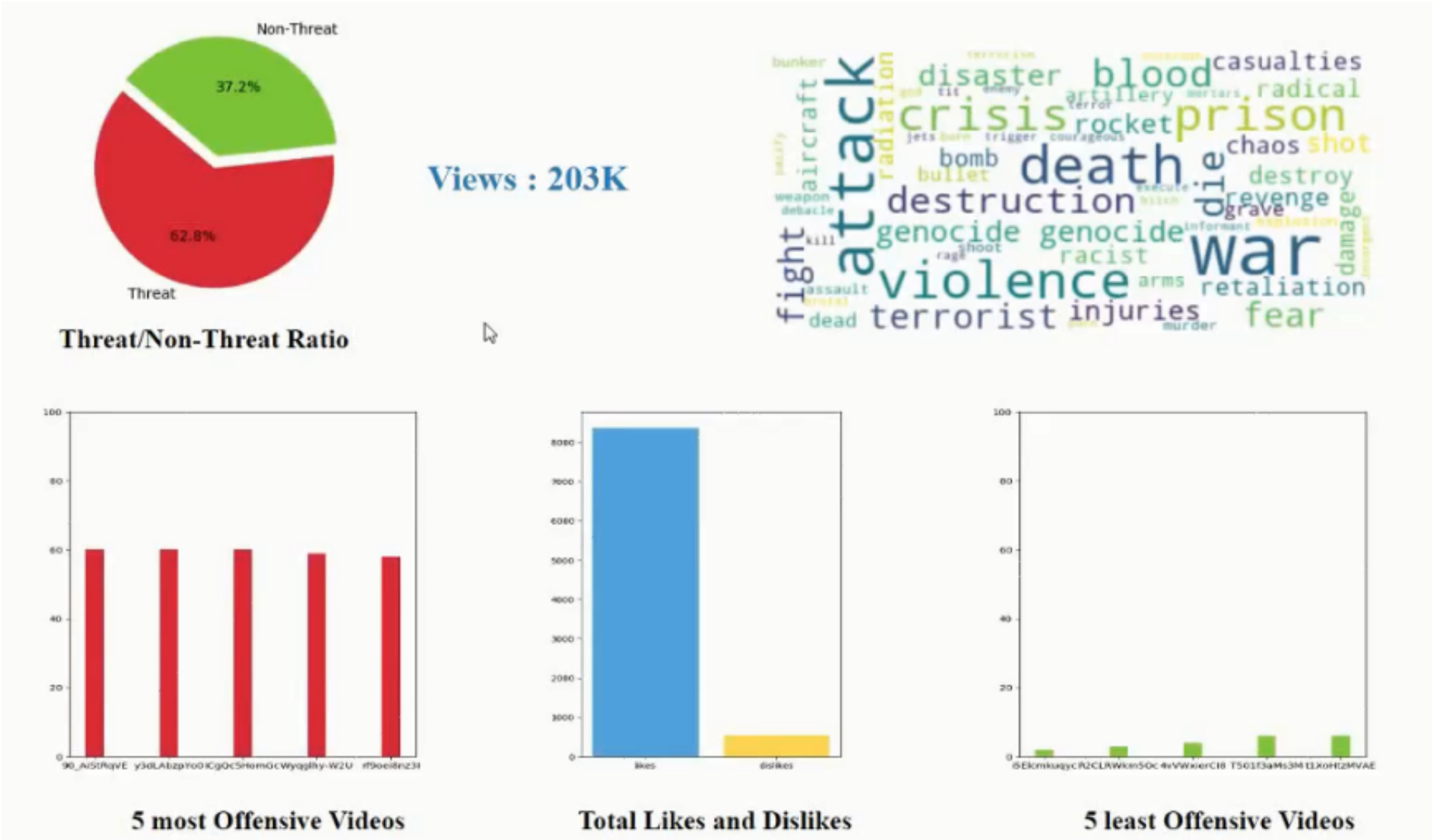

Figure 5: Statistical page for a youtube channel.

## 4 Conclusion

In this paper, we proposed a framework to identify videos that offend large groups of people in a community. We collected 20k+ videos based on keywords and extracted out all the key information available. We experimented with various combinations of features such as sentiments from comments, thumbnail images, text based features, video and frame level features. We have used state of the art Deep Learning models and embeddings to generate rich feature set for better analysis. We have a ready to use api which could easily be used to filter offensive youtube channels. We have also tried to provide a customisation in terms of types of offensive nature of videos that the user wants to filter out. This could help in scalability of the systems and could also be useful for different clients and customers since it can very well adapt to their needs.

One of our limitations of our approach is the limited dataset. The amount of data that we used to classify the videos are very less in comparison to the amount of data uploaded on youtube. To keep in par with such a massive data generation, our system should be embedded in youtube core framework so that it can update the learning and can be able to classify better. Another limitation is the restriction of language. We restricted our framework to only English videos. We can extend this framework to support multilingual content metadata. Also, computation of other features for extracting audio transcript associated with the video as text and further analyzing it, topic-modeling of the comments may improve performance of the classifier.

## References

1. [1] Y. Chen, Y. Zhou, S. Zhu, and H. Xu, “Detecting offensive language in social media to protect adolescent online safety,” in 2012 International Conference on Privacy, Security, Risk and Trust and 2012 International Conference on Social Computing, Sept 2012, pp. 71–80.
2. K. Dinakar, B. Jones, C. Havasi, H. Lieberman, and R. Picard, “Common sense reasoning for detection, prevention, and mitigation of cyberbullying,” ACM Trans. Interact. Intell. Syst., vol. 2, no. 3, pp. 18:1–18:30, Sep. 2012.
3. M. Dadvar, D. Trieschnigg, R. Ordelman, and F. de Jong, Improving Cyberbullying Detection with User Context. Berlin, Heidelberg: Springer Berlin Heidelberg, 2013, pp. 693–696.
4. K. Reynolds, A. Kontostathis, and L. Edwards, “Using machine learning to detect cyberbullying,” in 2011 10th International Conference on Machine Learning and Applications and Workshops, vol. 2, Dec 2011, pp. 241–244.
5. N. Aggarwal, S. Agrawal, and A. Sureka, “Mining youtube metadata for detecting privacy invading harassment and misdemeanor videos,” in 2014 Twelfth Annual International Conference on Privacy, Security and Trust, July 2014, pp. 84–93.
6. R. I. Rafiq, H. Hosseinmardi, R. Han, Q. Lv, S. Mishra, and S. A. Mattson, “Careful what you share in six seconds: Detecting cyberbullying instances in vine,” in 2015 IEEE/ACM International Conference on Advances in Social Networks Analysis and Mining (ASONAM), Aug 2015, pp. 617–622.
7. T. Fu, C. N. Huang, and H. Chen, “Identification of extremist videos in online video sharing sites,” in 2009 IEEE International Conference on Intelligence and Security Informatics, June 2009, pp. 179–181.